\documentclass[runningheads]{llncs}
\usepackage{graphicx}
\usepackage{amsmath, amssymb,mathtools}
\usepackage{amsfonts}
\usepackage{textcomp}
\usepackage{subfig}
\usepackage{booktabs, algorithm, algorithmic, multirow}
\usepackage[utf8]{inputenc}
\usepackage[english]{babel}
\usepackage{csquotes}
\usepackage{subfig}
\usepackage{bibnames}

\usepackage{array}

\usepackage{url}

\usepackage{breakurl}
\usepackage[breaklinks]{hyperref}

\begin{document}
\title{Domino Saliency Metrics: Improving Existing Channel Saliency Metrics with Structural Information}
\titlerunning{Domino Saliency Metrics}
\author{Kaveena Persand
\email{persandk@tcd.ie}
\and Andrew Anderson
\email{aanderso@tcd.ie}
\and David Gregg
\email{david.gregg@cs.tcd.ie}}

\authorrunning{Persand et al.}
\institute{Trinity College Dublin, Dublin 2, Ireland}
\maketitle              % typeset the header of the contribution

\begin{abstract}

Channel pruning is used to reduce the number of weights in a Convolutional
Neural Network (CNN).  Channel pruning removes slices of the weight tensor so
that the convolution layer remains dense.  The removal of these weight slices
from a single layer causes mismatching number of feature maps between layers of
the network.  A simple solution is to force the number of feature map between
layers to match through the removal of weight slices from subsequent layers.
This additional constraint becomes more apparent in DNNs with branches where
multiple channels need to be pruned together to keep the network dense.
Popular pruning saliency metrics do not factor in the structural dependencies
that arise in DNNs with branches.  We propose Domino metrics (built on existing
channel saliency metrics) to reflect these structural constraints.   We test
Domino saliency metrics against the baseline channel saliency metrics on
multiple networks with branches.  Domino saliency metrics improved pruning
rates in most tested networks and up to 25\% in AlexNet on CIFAR-10.


\end{abstract}

\section{Introduction}
Deep neural networks can reach human level accuracy for many
classfication problems~\cite{TaigmanYRW14}, but they have huge memory
and computation cost.  Pruning reduces the size of neural networks via
the removal of unnecessary weights\cite{Mozer_Smolensky,LWC}.  Channel
pruning removes weights corresponding to an entire channel in the
output of a layer. When the weights of the entire channel
are set to zero, the corresponding output feature map of the channel
becomes zero. This zero output feature map feeds into subsequent layers
which may allow the corresponding channel to be removed from these
subsequent layers in a \textit{domino effect}.

In a network architecture where each layer has exactly one output and one input
layer, the removal of an output feature map leads to the following layer's
input feature map no longer contributing to the network.  This scenario is
illustrated in Figure \ref{fig:pruning_base} where the consumer layer
is a convolution or fully-connected layer, the resulting zero input
channel allows the corresponding weights from the consumer layer
to be pruned.

\begin{figure}
\centering
\includegraphics[width=0.99\linewidth]{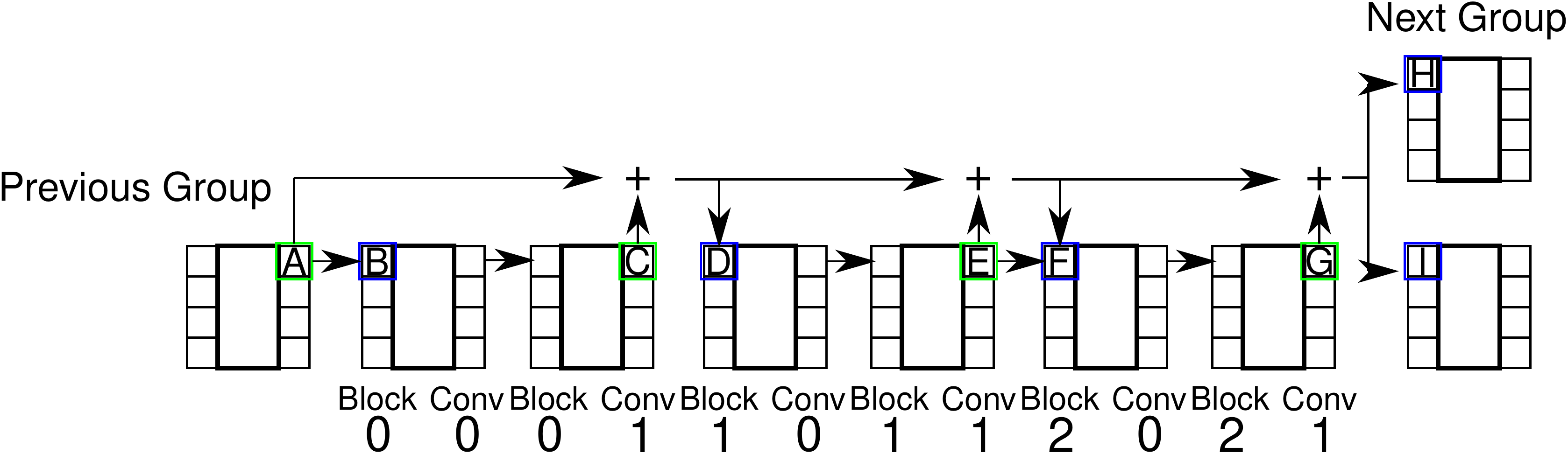}
\caption{Structure of a block in ResNet-20.  The direction of arrows show the
direction in which data flows.}
\label{fig:resnet_data}
\end{figure}

In networks with branches/splits, the output feature map from one layer may
feed into multiple others. With joins, feature maps from different layers feed
into a single layer. A common occurence of split and join connections in CNNs
is due to skip connections, which were first pioneered in ResNet architectures.
The common structure of a ResNet block containing join and split connections is
shown in Figure \ref{fig:resnet_data}.  The presence of these joins allows
multiple output feature maps to be removed together when one feature map is
considered for removal.  Networks that offer state-of-the-art accuracy for
image classification often contain skip
connections~\cite{nfnet-paper,resnest-paper,resnext-paper,Mahajan2018,Touvron2019}.
ResNet architectures are also more difficult to prune for their lower
redundancy ~\cite{Thinet2017,MoreIsLess,He2017}.  Hence, improving pruning
rates for networks containing skip connections can be very advantageous.

Another common occurence of joins is
group convolution.  Group convolution was originally used in the AlexNet
architecture ~\cite{alexnet-paper} to parallelise convolution on multiple GPUs.
Since, it has also been used in state of the art architectures
~\cite{resnext-paper,nfnet-paper}.

Most approaches do not factor in the removal of different feature maps
when computing the saliency metric used for pruning.  We propose using
\textit{Domino saliency metrics} to factor in the saliency of feature
maps and weights that need to removed together. We make the following
contributions: (1) We propose Domino saliency metrics, where existing
saliency metrics are combined together depending on the set of
channels that need to be removed together. (2) Combining channel
saliency metrics to obtain Domino saliency metrics, has a negligible
computational cost to computing channel saliency metrics and requires
very little modification to existing pruning strategies. (3) We
experimentally evaluate two variants of Domino saliency metrics:
$Domino-o$ and $Domino-io$, and find that they significantly improve
pruning.

%The removal of a feature map from the network not only
% reduces the intermittent memory and computational cost at the pruned
% layer but also for following layers.

%The common structure of a convolutional neural network used for classification
%can be broken down into two parts: the feature extraction layers, and the
%classification layers.  The feature extraction layers are typically an
%arrangement of convolution layers and activation layers.  The classification
%layers are either a succession of fully-connected layers followed by activation
%layers or an activation layer.

\section{Data Flow Graph for Pruning}

\subsection{Background}

Pruning is the removal of weights from the network. The pruned weights are set
to zero.  Pruning can be done in an unstructured or structured way.
Unstructured pruning removes weights from the network without any given
constraint in pattern.  On the other hand, structured pruning removes weights
in a chosen pattern.  Common patterns for structured pruning are (in increasing
size of pattern): intra-kernel, kernel, and channels (or filters).

Channel pruning removes entire channels from convolution layers of the network.
By removing entire channels, the weight tensor remains dense, so existing DNN
dense libraries can be used. The removal of an entire channel of convolution
weights  leads to the removal of its subsequent feature map. Removing a 
feature map may allow corresponding weights from subsequent layers to be
removed. In a network where each layer has at most one input layer and output
layer, this relationship is obvious.

\subsection{Channel Pruning Networks with Splits and Joins}

While simpler neural networks are often linear and acyclic
\cite{lenet-paper}\cite{cifar10-paper}, modern networks often contain join
nodes and split nodes\cite{alexnet-paper,resnet-paper}.  

It is obvious which weights need to be removed when applying channel pruning to
a network where each layer is fed to only one successor.  However in networks
with branches, feature maps are used by more than one layer.  A layer can have
multiple successors due to skip connections.  Skip connections are elementwise
additions between output feature maps of different convolutions to produce the
input feature map of the following layer.  

In networks with branches, applying a reachability analysis such as is
performed by a compiler on a more traditional computational structure, the
control-flow graph, to uncover the weights that need to be pruned together to
keep the network dense.  The basic intuition here is that if the definition of
some feature map \emph{reaches} a layer, the pruning of that feature map may also
imply the removal of more feature maps which are computed from it. While this
seems trivial for linear networks, the introduction of splits and joins in the
graph mean that extra care must be taken in order to exploit the dependence
relationship to achieve better pruning results.

\subsection{Data Flow Graph}

Neural networks form a directed graph structure where simple input-output dependence
exists between producer and consumer layers in the network (i.e. the network
forms a \emph{data flow graph}).  When representing the data flow graph, we can
make abstraction of activation layers (see Section
\ref{ssec:prune_activation}).  Hence, we only need to represent the data flow
between a convolution or fully-connected layer to other convolution or
fully-connected layers.  

We introduce some notation to facilitate the description of the data flow
graph.  A layer $l$ has a set of 3-dimensional input feature maps $I(l)$, a set
of 3-dimensional output feature maps $O(l)$ and weights $W^l$ (with shape
$m^l_{out} \times m^l_{in} \times k^l \times k^l$).  $O_i(l)$ and $I_i(l)$
refer to the $i^{th}$ 2-dimensional feature map of $O(l)$ and $I(l)$ respectively.
We can describe the flow of information between layers with a successor
relation $O(l) = I(succ(l))$.  Intuitively, the \emph{data flow successors}
$succ(l)$ of a layer $l$ are those layers whose input is the output of layer
$l$.

When a layer in a neural network joins the output of multiple producer
channels, we write $O(l) \subset I(succ(l))$ i.e. the successor relation
extends to any channel which consumes the output of  $l$. If $I(l+1)$ is the
input feature map of the following, then it is the direct successor of $O(l)$.

\subsection{Join and Split Nodes}

\begin{figure}
\centering
\subfloat
[Join connection in CNN.]{\label{fig:pruning_join}\includegraphics[width=0.48\linewidth]{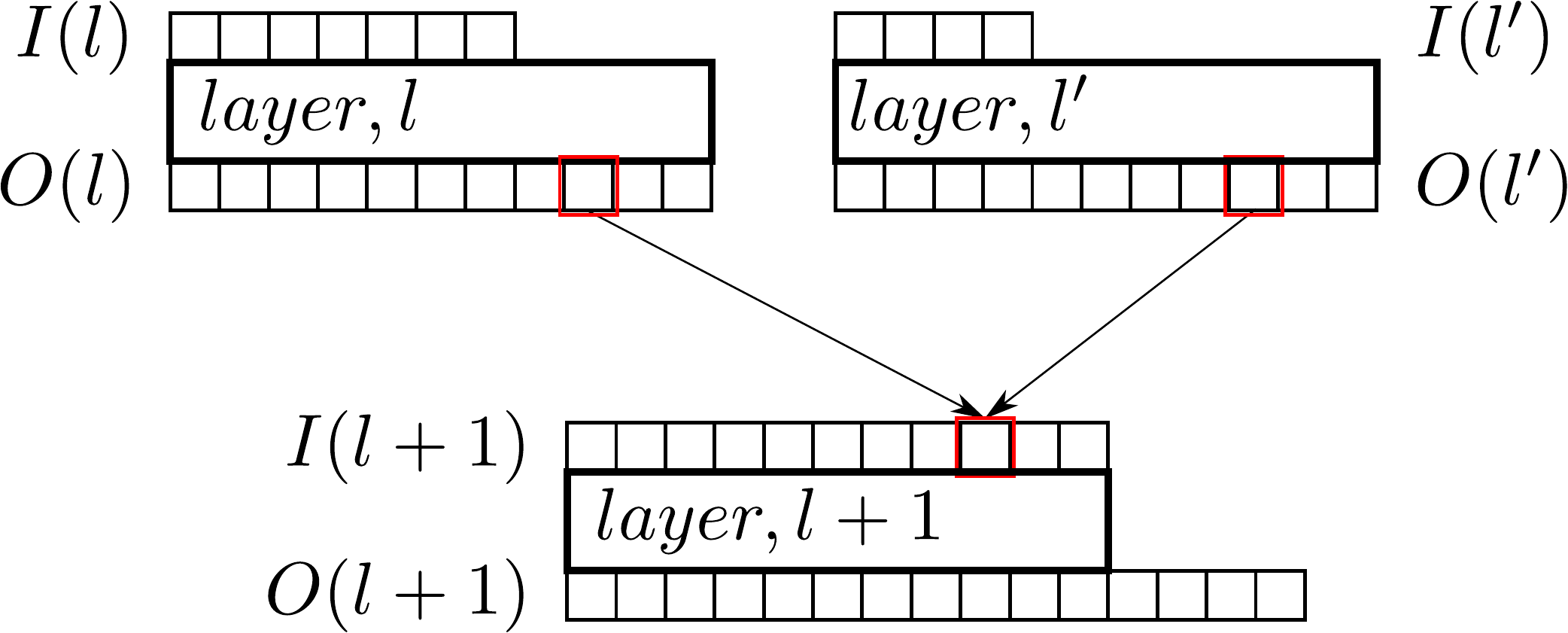}}
\subfloat
[Split connection in CNN.]{\label{fig:pruning_split}\includegraphics[width=0.48\linewidth]{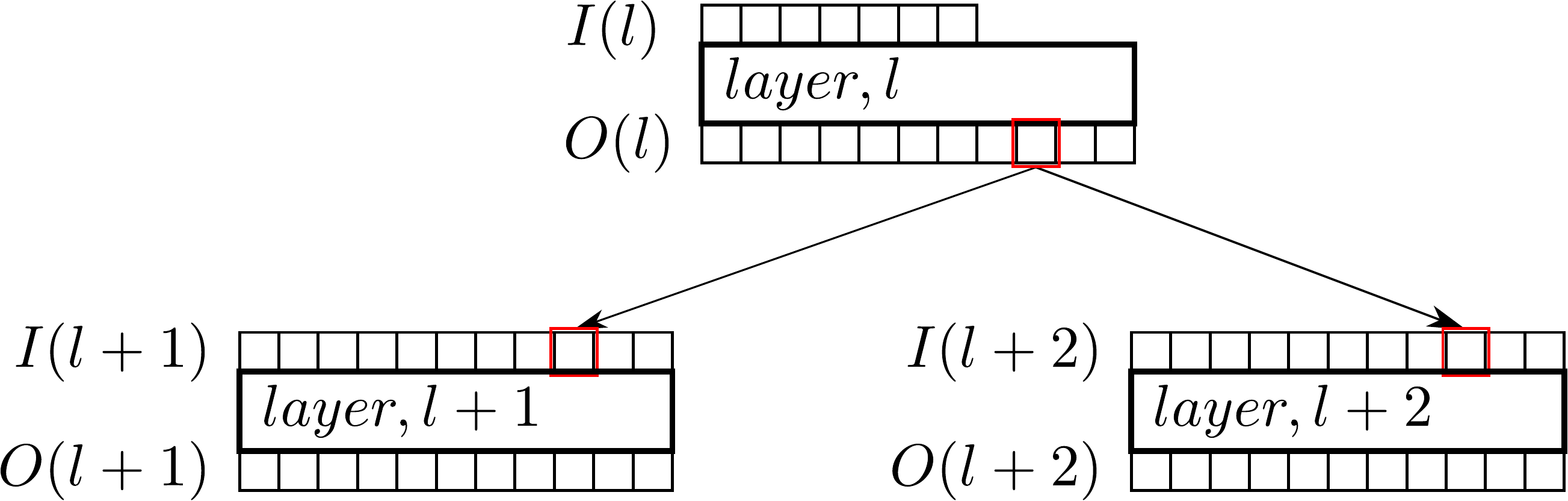}}
\caption{Data dependencies with join and split connections in CNNs. The arrows are in the direction of data flow. }
\label{fig:join_split}
\end{figure}

$I(l+1)$ is not necessarily the only successor of $O(l)$.  For example, in
Figure \ref{fig:pruning_join} the elementwise sum of two layers $l$ and $l'$ is
fed to layer $l+1$. $l+1$ satisfies $O(l) \subset I(l+1) \land O(l') \subset
I(l+1)$.  We can also consider data dependencies for a single channel, $C^l_i$, instead
of an entire layer, $l$, with $O(C^l_i)$ = $O_i(l)$.  So we would say
that both $succ(C^{l}_i) = \{C^{l+1}_i\}$ and $succ(C^{l'}_i) = \{C^{l+1}_i\}$.
Split, or broadcast relationships also exist, where the output of a layer is
used by multiple consumers. An example of split in the data flow graph is seen
in Figure \ref{fig:pruning_split} where $succ(C^{l}_i) = \{C^{l+1}_i,
C^{l+2}_i\}$. 

\subsection{Group Convolution}

\begin{figure}
\centering
\subfloat
[Group convolution with $g^l = 2, m^{l}_{in} = 2, m^{l-1}_{out} = 4$.]{\label{fig:pruning_group_op}\includegraphics[width=0.7\linewidth]{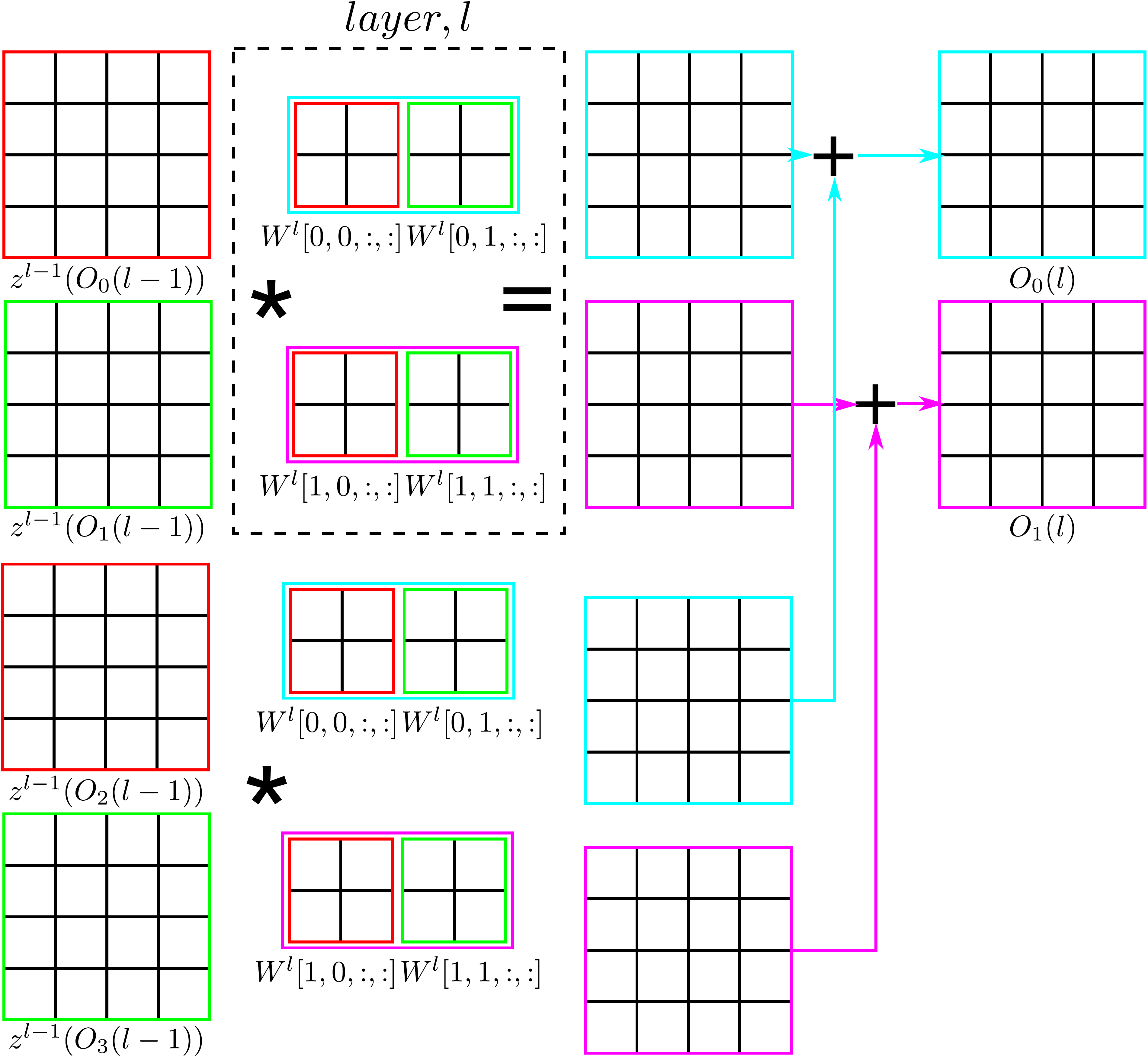}}
\subfloat
[Group convolution data dependency.]{\label{fig:pruning_group}\includegraphics[width=0.25\linewidth]{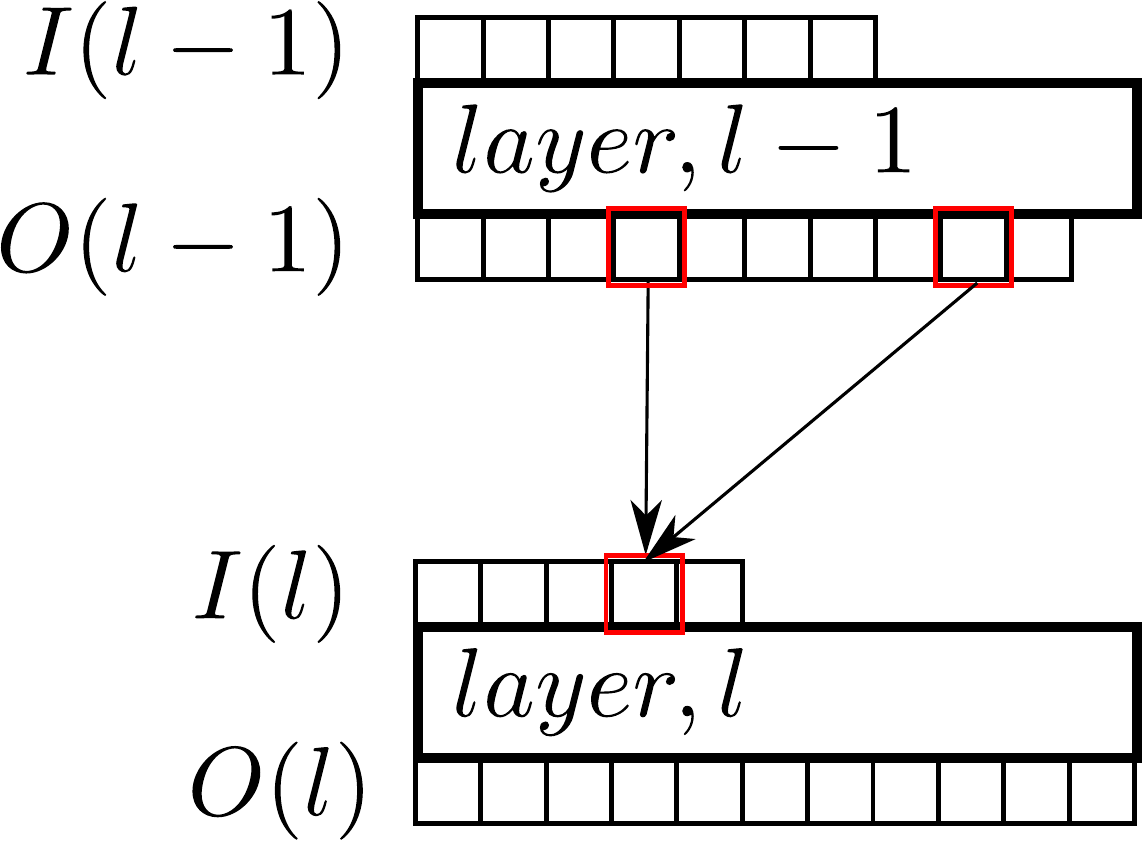}}
\caption{Data flow with group convolutions in CNNs. The arrows are in the direction of data flow.}
\label{fig:data_group}
\end{figure}

Data dependencies at the entrance of group convolutions can be modeled using a
simple convolution layer that has $g$ sets of input feature maps that result in
$g$ sets of partial output feature maps.  These partial output feature maps are
then added together to create the final output feature map of the layer.  We
have $m^{l+1}_{in} = \frac{m^{l}_{out}}{g^l}$ with the group size, $g^l > 1$.
Figure \ref{fig:data_group} illustrates a group convolution with a group size
of 2.

Hence, $O(C^{l-1}_i) \subset I(C^{l}_i) \land O(C^{l-1}_{i + g^l}) \subset
I(C^{l}_i)$.  In Figure \ref{fig:pruning_group}, $succ(C^{l-1}_i) = \{
C^{l}_{i}, C^{l}_{i + 2} \}$.

Considering a neural network graph with $G$ channels, we can define the successor
relation as:

\begin{equation}
    \forall c\in G, \exists x \in G : O(c) \subset I(x) \Rightarrow x\in succ(c) \tag{\textsc{successor}}\label{eqn:succ}
\end{equation}

Let the predicate $P(c)$ be true where a feature map $c$ is being
pruned, and false otherwise. The truth of $P(c)$ is determined locally for the
input or output feature maps of a specific layer by the pruning process.

We are interested in how the truth of $P(c)$ locally in any single layer may
influence the truth of $P(c)$ for connected layers in the network, or more
informally, how pruning some feature maps  may imply the pruning of other
feature maps.

\begin{equation} \forall c \in G, \forall x\in succ(c) : P(O(c)) \Leftrightarrow P(I(x))
\tag{\textsc{Channel Pruning}}\label{eqn:channel_pruning} \end{equation}

Equation \ref{eqn:channel_pruning} states that the pruning an output feature
map $O(c)$ is materially equivalent to pruning the input feature map $I(x)$,
with $x$ a successor of $c$.  This predicate is valid for all the channels in
the network.  Hence, in a typical ResNet style block this leads to the
simultaneous pruning of multiple channels.  In Figure \ref{fig:resnet_data},
feature maps A-I need to be removed simultaneously.

Equation \ref{eqn:channel_pruning} is used to propagate pruning of feature maps.  However, for channel pruning
we need to remove weights from the network.  To prune an output channel,
$c=C^l_i$, from the network, weights $W^l[i, :, :, :]$ are removed from network
with ``$:$" representing all valid indices.  Hence, when a feature map
$O(C^l_i)$ is pruned, $W^l[i, :, :, :]$ is pruned and when a feature map
$I(C^l_i)$ is pruned, $W^l[:, i, :, :]$ is pruned.  

\subsection{How to Prune Biases and Activation Layers}
\label{ssec:prune_activation}

The output of a convolution is not often fed directly to the next
convolution or fully-connected layer.  Instead, at least one activation
function is applied to the feature map before being fed to the next layer.
Hence, the output feature map produced by a layer is not directly equivalent to
the input feature map of the following layer.

For most activation functions, applying them to a zero feature map results in a
zero feature map.  ReLU, GELU, and max/average pooling layers, are examples of
activation functions that output zero for a zero input.  When these activation
functions are used, a pruned output feature map results in pruned input feature
maps of the subsequent layers.  However, in the case of activation functions with
biases such as Batch Norm or bias layers, a zero input does not always result
into a zero output unless the biases are also set to zero.  Hence, the pruned
feature map and the pruned weights cannot be removed from the network.
However, if the corresponding biases are set to zero, the weights can then be
removed from the network.

\begin{figure}[]
\centering
\includegraphics[width=\linewidth]{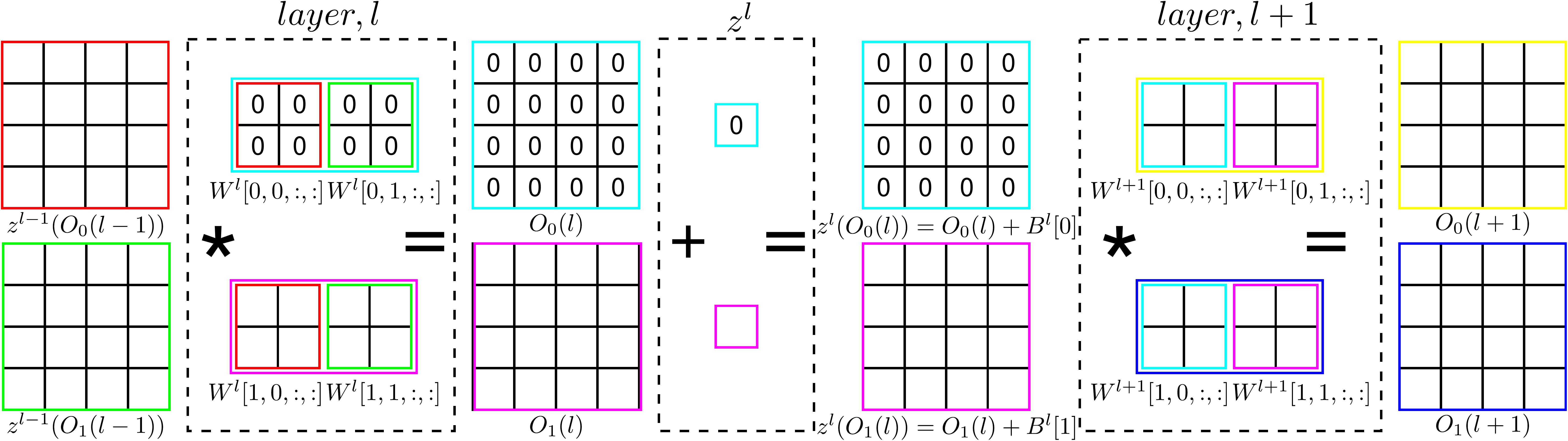}
\caption{Channel pruning where the outputs of convolution $l$ are fed into an
activation layer (with bias), $z^l$ before being fed to the successor $l+1$.}
\label{fig:pruning_base}
\end{figure}

A simple channel pruning case is shown in Figure \ref{fig:pruning_base},  For
the convolution layer $l$, to produce a single $O_i(l)$ each $j^{th}$ input
feature map is convolved with its corresponding 2D filter $W^{l}[i,j,:,:]$ and
summed together.  In Figure \ref{fig:pruning_base}, $O_0(l) = z^{l-1}(O_0(l-1))
* W^{l}[0,0,:,:] + z^{l-1}(O_1(l-1)) * W^{l}[0,1,:,:]$ and $O_1(l)=
z^{l-1}(O_0(l-1)) * W^{l}[1,0,:,:] + z^{l-1}(O_1(l-1)) * W^{l}[1,1,:,:]$.  To
prune the output channel $C^{l}_0$, $W^{l}[0,0,:,:]$ and $W^{l}[0,1,:,:]$ are
set to zero.  If the corresponding bias in the activation layer $z^l$
($B^l_0$), is set to zero, then the input feature map $z^l(O_0(l))$ is also
zero.  Since convolution with a zero feature map results into a zero feature
map, the values of $W^{l+1}[0,0,:,:]$ and $W^{l+1}[1,0,:,:]$ no longer
influence feature maps $O_0(l+1)$ and $O_1(l+1)$.  Hence, $W^{l}[0,0,:,:]$,
$W^{l}[0,1,:,:]$, $B^l_0$, $W^{l+1}[0,0,:,:]$, and $W^{l+1}[1,0,:,:]$, can be
set to zero and removed.

If the biases are not set to zero, then feature maps filled with zeroes still
need to be stored to keep the network dense.  In practice, output feature maps
filled with zeroes are removed from the network to increase memory savings.
Hence, biases of activation functions are also set to zero and the obsolete
parameters are removed to keep the network dense.  Hence, to simplify the data
flow graph, we can make abstraction of activation functions even if they
contain biases.  We can also assume that when a channel is pruned, weights,
feature maps and activation layer parameters of that channel are set to zero.

\section{Domino Pruning}

When pruning neural networks with skip or group connections, most approaches
use the same saliency metrics as for simple forward feed neural networks
without any modification.  The construction of these saliency metrics do not
take into account the structural dependencies that may to be satisfied between
layers to keep the network dense. 

We argue that a more straightforward inspection of the data dependence
structure may be more prudent. Using a reachability analysis such as may be
performed by a compiler, we show how the removal of one set of parameter may be used
to heuristically perform a cascade of removals of other reachable parameters.
We refer to this technique as ``Domino" saliency metrics.  Any saliency metric
that has been formulated to give each channel a saliency measure can be used
with Domino saliency metrics.  

\begin{equation} \forall x\in succ(c), \exists y : x \in succ(y) \Rightarrow y
\in coparent(c) \tag{\textsc{coparent}}\label{eqn:coparent} \end{equation}

\begin{equation} 
\forall x \in succ(c), \exists y: y \in succ(c) \Rightarrow y \in sibling(x)
\tag{\textsc{sibling}}
\label{eqn:sibling} \end{equation}

With Domino pruning, the output channels that are considered coparents are
considered as a single pruning choice.  Hence, they cannot be removed without
the other coparent output channels.  Output channels are considered coparent
if they share a common successor as shown in Equation \ref{eqn:coparent}.
Similarly Equation \ref{eqn:sibling} is used to find siblings of direct
successors.  We consider the $coparent$ and $sibling$ relationships to be
birectional and transitive, i.e., if $A \in coparent(B)$ then $B \in
coparent(A)$ and if, $C \in coparent(B)$ then $C \in coparent(A)$. $coparents^+$
and $siblings^+$ are the transitive closure of the $coparent$ and $sibling$
relationships.  
When $c$ is an output channel, then $siblings^+(c)$ represents the transitive closure of any of $c$'s successors and when $c$ is an input channel, $coparents^+(c)$ represents the transitive closure of any output channel that is used to produce $c$. 
The feature maps and weights removed when one output channel is
removed is given by Equation
\ref{eqn:channel_pruning}.  The set of parameters $coparents^+(c) \cup siblings^+(c)$ is
pruned when the output channel $c$ is pruned.  In the case of a network with
no joins and no splits $coparents^+(c) = \{ c \}$ and $siblings^+ = succ(c)$.

Saliency metrics are traditionally computed for a single output channel.  This
is also true for networks with skip connections (ResNets).  With Domino metrics
we combine the channel saliency of channels that are removed together.

We propose two variants of Domino saliency: $Domino-o$ and $Domino-io$.

$Domino-o$ adds the saliency of all output channels that are removed together.
Since most channel pruning algorithms use the channel saliency of output
channels, $Domino-o$ has negligible cost and require minimal change to the
pruning algorithm.  Equation \ref{eqn:domino_saliency_o} decribes how to
compute $Domino-o$ for a channel using $coparents^+$ and their channel saliency
$S$.

\begin{equation} Domino(c) = \sum_{x \in coparents^+(c)} S(x)
\label{eqn:domino_saliency_o}\end{equation}

As illustrated in Figure \ref{fig:pruning_base}, the output feature produced by
a convolution channel ultimately becomes the input feature map for the
following layer.  Few approaches apart from feature reconstruction based
metrics, exploit this relationship for saliency computation.  With $Domino-io$,
we add the saliency of all the weights or feature maps that are removed when a
channel is removed to get the saliency of the channel to be pruned.  Equation
\ref{eqn:domino_saliency_io} shows how to compute $Domino-io$ using the channel
saliency of $coparents^+$ (output feature maps or weights) and $siblings^+$
(input feature maps or weights) using a channel saliency $S$.

\begin{equation} Domino-io(c) = \sum_{x \in coparents^+(c)} S(x) + \sum_{x \in
siblings^+(c)} S(x) \label{eqn:domino_saliency_io}\end{equation}

%Figure \ref{fig:domino_saliency} shows an example of how Domino saliency metrics are computed.
%
%\begin{figure}
%\centering
%\includegraphics[width=0.7\linewidth]{figures/domino_saliency.pdf}
%\caption{Domino saliency metrics.}
%\label{fig:domino_saliency}
%\end{figure}

\section{Experimental Evaluation}

Channel saliency is the use of the saliency metric of a single output channel
to prune all the dependent weights and feature maps.  This is the common
strategy when pruning networks.  In practice, this leads to the lowest saliency
of the output channels that are pruned together to be used as saliency metric
of the set of weights or feature maps to be pruned.
This baseline is denoted $S$.

We compare Domino saliency metrics constructed using a channel saliency against
the use of the base channel saliency.

A Domino saliency metric is built using a baseline channel saliency
metric.  If the saliency of all channels are roughly of the same order
and positive, their addition is of greater order than channels that are not part of
split or join nodes.  We use an average of the number of weights or output
points to avoid favoring isolated channels for pruning.  Saliency metrics that
use feature maps are scaled using the number of pixels in the feature maps.
For example, if two saliency $S(c)$ and $S(z)$ are computed using $N_c$ and
$N_z$ weights then, the Domino-scaled metric is $\frac{S(c) + S(z)}{N_c + N_z}$
instead of $S(c) + S(z)$.  The scaled channel metrics are $\frac{S(c)}{N_c}$
and $\frac{S(z)}{N_z}$.  Metric with this average is suffixed with -avg.

\subsection{Pruning Algorithm}

The saliency metric is one component of the pruning algorithm.  We choose a
pruning algorithm that heavily relies on the choices of the saliency metric to
determine the improvement brought by changing the saliency metric.  Since our aim is to
find better saliency metrics, we avoid obfuscating the contribution of the
saliency metric by not retraining after each pruning step.  If a saliency
metric is able to achieve higher pruning rates without retraining, then it can
also be used to reduce the cost of retraining\cite{taxonomy-paper}.

For each pruning iteration we compute the saliency of every output channel.  The
lowest saliency channel is then pruned according to Equation
\ref{eqn:channel_pruning}.  This process is repeated until the test accuracy
falls under 5\% of its initial value.

\subsection{Networks}
\label{ssec:networks}

We evaluate Domino saliency metrics on popular architectures with split and
join nodes.  We evaluate the original ResNet architecture~\cite{resnet-paper}
and a state-of-the-art ResNet-inspired architecture,
NFNET-F0~\cite{nfnet-paper}.  We also evaluate Domino saliency metrics on
AlexNet~\cite{alexnet-paper}.  The join and split connections in ResNet arise
due to skip connections.  The join connections in AlexNet arise due to group
convolutions.  NFNET-F0 contain both skip connections and group convolutions.

We use the CIFAR-10\cite{cifar10-paper}, CIFAR-100\cite{cifar10-paper},
ImageNet-32\cite{Chrabaszcz} (a downsized ImageNet variant), and ImageNet\cite{imagenet-paper} datasets.  ResNet-20 \footnote{https://github.com/HolmesShuan/ResNet-18-Caffemodel-on-ImageNet} on ImageNet, ResNet-50\cite{pretrained-resnet50} on ImageNet and NFNET-F0~\cite{nfnet-paper} on ImageNet are pretrained networks.
The remaining networks are trained from scratch.

\subsection{Saliency Metrics}

The most popular pruning saliency metrics are either a derivative of the L1 or
L2 norm of
weights~\cite{LWC,DSD,AMC,Frankle_Carbin,DynamicNetworkSurgery,Dally,Li2017,StructuredProbabilisticPruning,GroupWiseBrainDamage,SoftFilterPruning}
or Taylor
expansions~\cite{Lecun,Hassibi,Molchanov,Molchanov2019,CollaborativeChannelPruning,GlobalSparseMomentum}.
With channel pruning, Taylor Taylor expansions can be applied to either weights
or feature maps.  

The channel saliency metrics that we evaluate are: Taylor expansion using weights,  Taylor expansion using feature maps, and L1 norm of weights.

\section{Results}
\label{sec:results}

We measure the improvement of Domino saliency metrics by comparing the
percentage of convolution weights removed for a drop of 5\% in test accuracy.
The results are an average from 4 runs.

\begin{figure}
\centering
\includegraphics[width=0.99\linewidth]{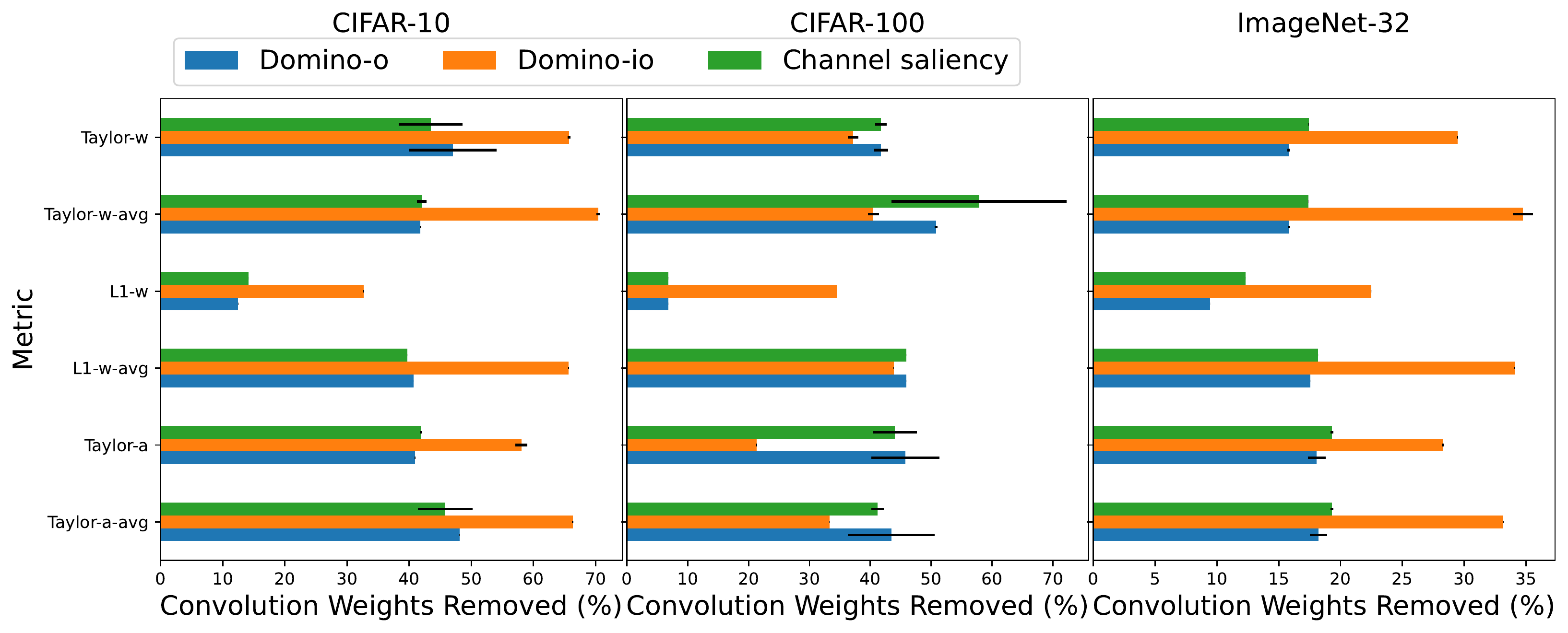}
\caption{AlexNet}
\label{fig:results_alexnet}
\end{figure}

\begin{figure}[h]
\centering
\includegraphics[width=0.99\linewidth]{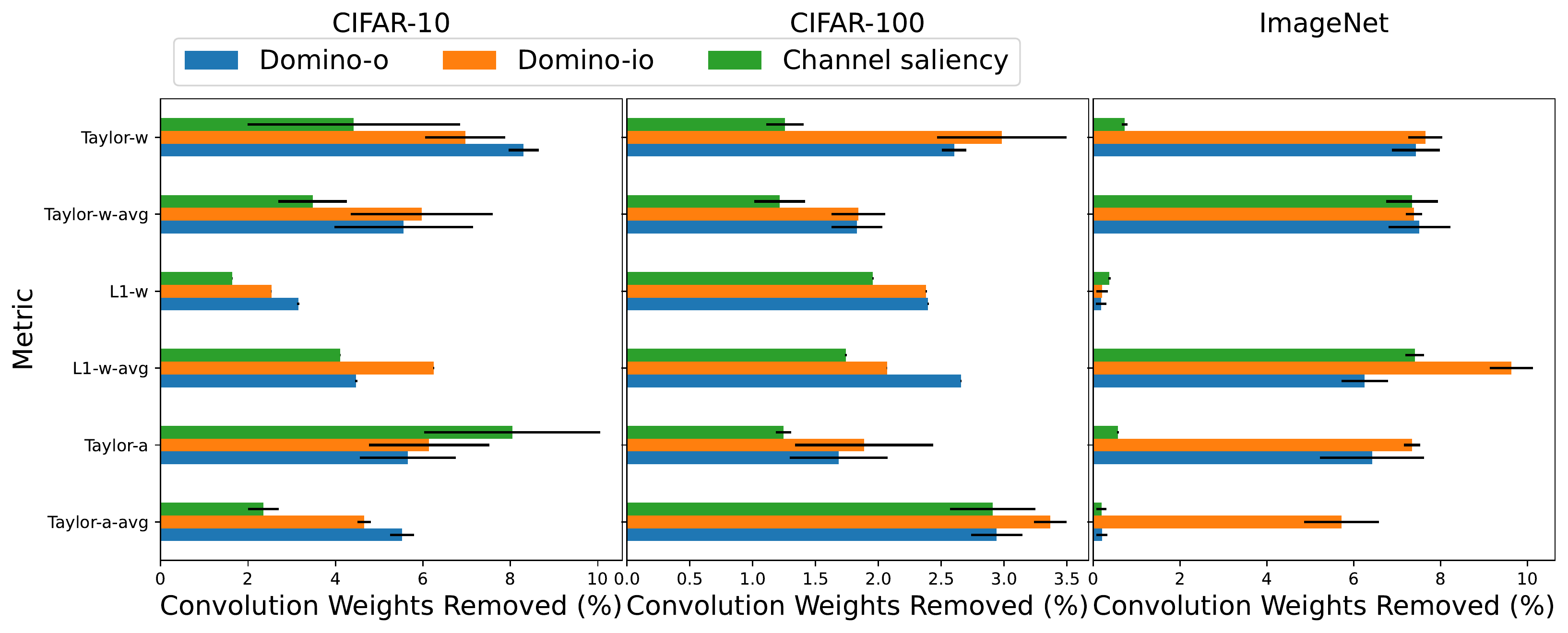}
\caption{ResNet-20}
\label{fig:results_resnet20}
\end{figure}

\begin{figure}[h]
\centering
\includegraphics[width=0.72\linewidth]{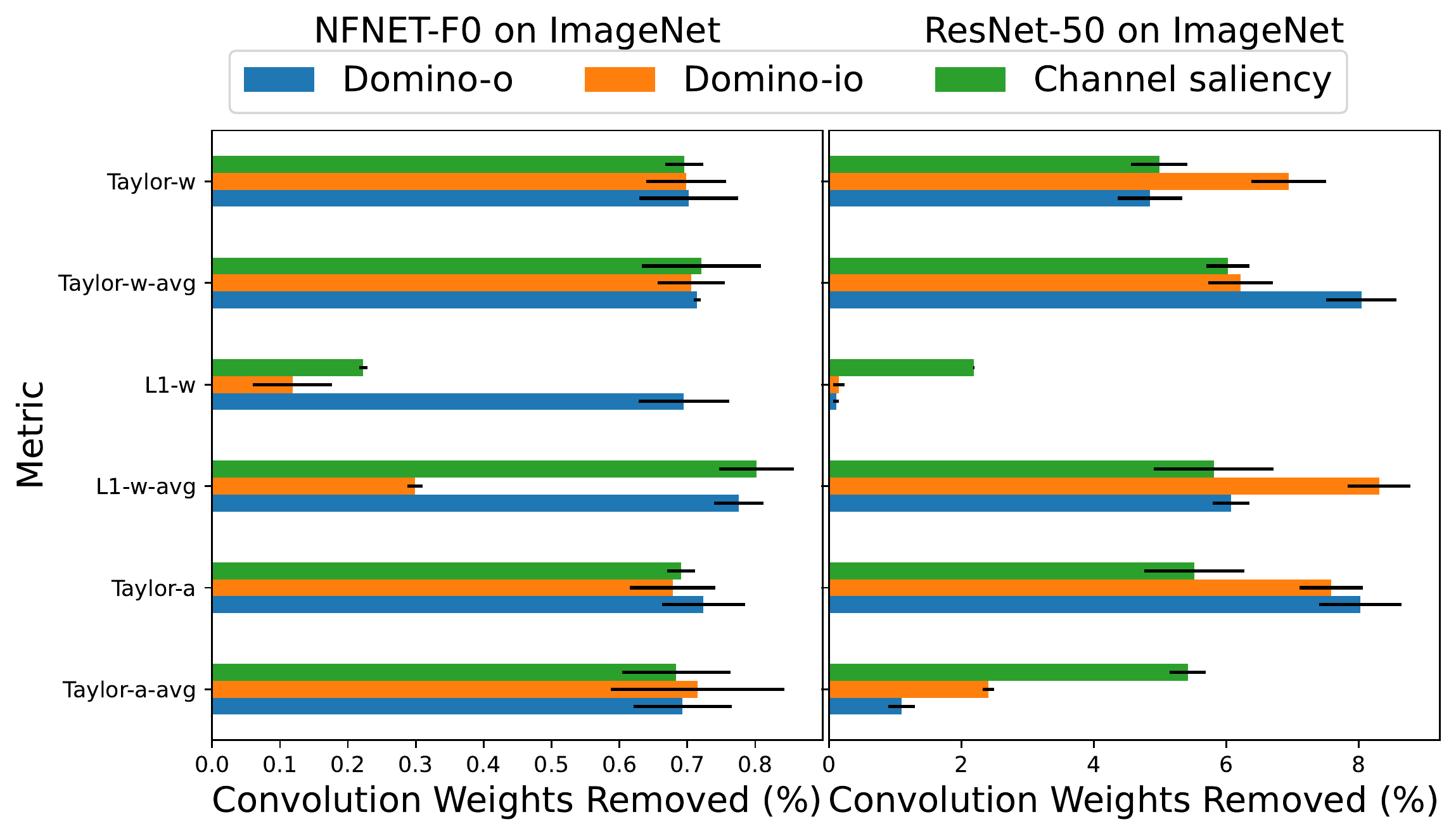}
\caption{NFNET-F0 and ResNet-50}
\label{fig:results_resnet50_nfnet}
\end{figure}

Figures \ref{fig:results_alexnet}, \ref{fig:results_resnet20}, and
\ref{fig:results_resnet50_nfnet} show the percentage of convolution
weights removed.  In most cases, we observe an improvement by using the Domino
saliency metrics.  $Domino-io$ significantly improves the pruning rates for
AlexNet on CIFAR-10, AlexNet on ImageNet-32 and ResNet-20 on CIFAR-10.
$Domino-io$  includes the saliency of input feature maps or input channel
weights in addition to $Domino-o$.  The larger improvement brought by
$Domino-io$ suggests that the saliency of input feature maps or input channel
weights contain relevant information for pruning.

From Figure \ref{fig:results_alexnet}, we see that using $Domino-io$ on AlexNet
for CIFAR-10 and ImageNet greatly improves the base saliency metric.  A notable
result is the L1 norm of weights (with averaging) which can match the pruning
rates of Taylor expansion based methods with $Domino-io$.  The L1 norm of
weights is a very popular metric for pruning for its low computational cost and
good pruning rates.  $Domino-io$ has a negligible cost overhead while greatly
improving the L1 norm of weights (with averaging).

From Figure \ref{fig:results_resnet20}, we observe a similar trend where the L1
norm of weights (with averaging) can be improved to match and exceed the
pruning rates of Taylor based method on ResNet-20 on ImageNet.

Figure \ref{fig:results_resnet50_nfnet} shows that the improvement of Domino
saliency metrics are marginal on NFNET-F0. However since the pruning rates are
also extremely low, the results are inconclusive for this network.  On the
other hand, ResNet-50 benefits from the additional structural information used
by Domino saliency metrics.  The pruning rates on ResNet-50 can be improved by
a few percentage points with either $Domino-o$ or $Domino-io$.

\begin{figure}
\centering
\subfloat[Average improvement.]{\includegraphics[width=0.49\linewidth]{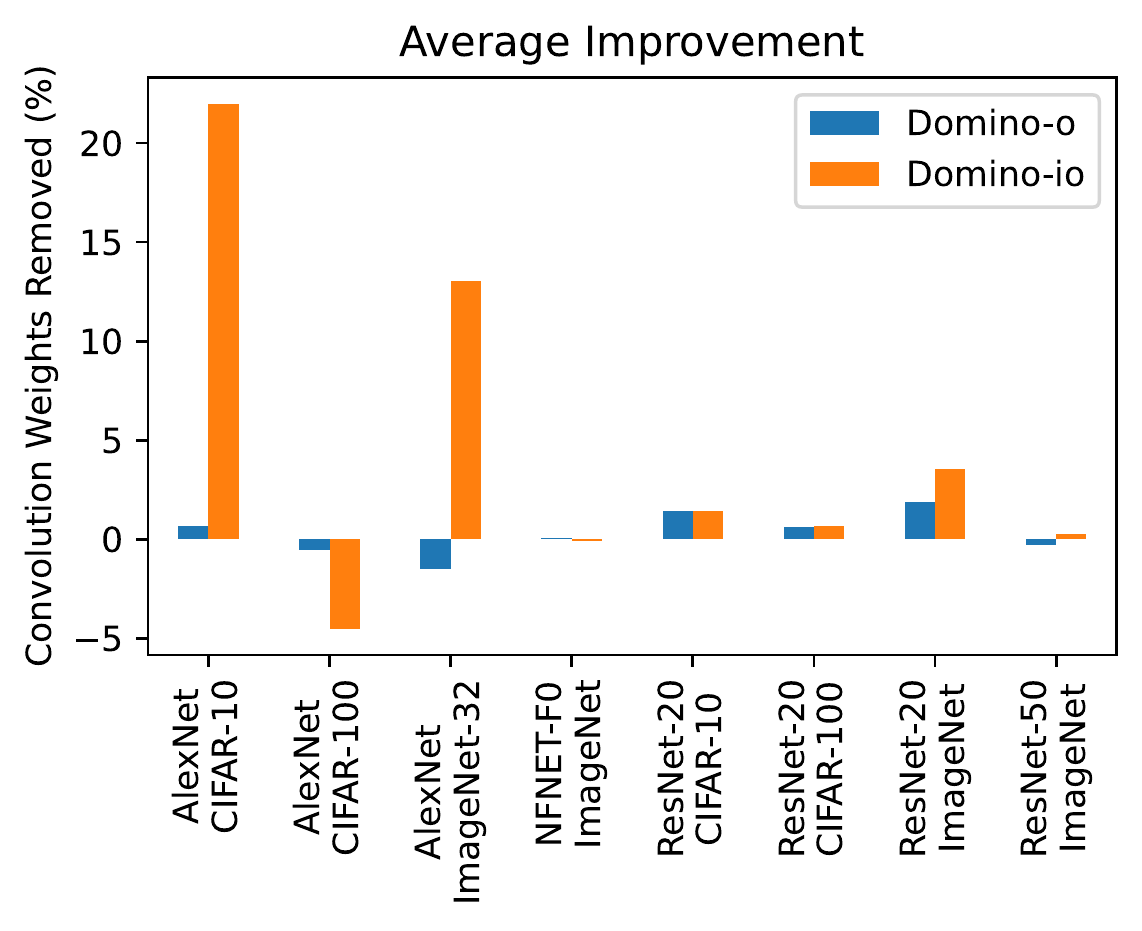}\label{fig:improvement}}
\subfloat[Improvement in maximum weights removed.]{\includegraphics[width=0.49\linewidth]{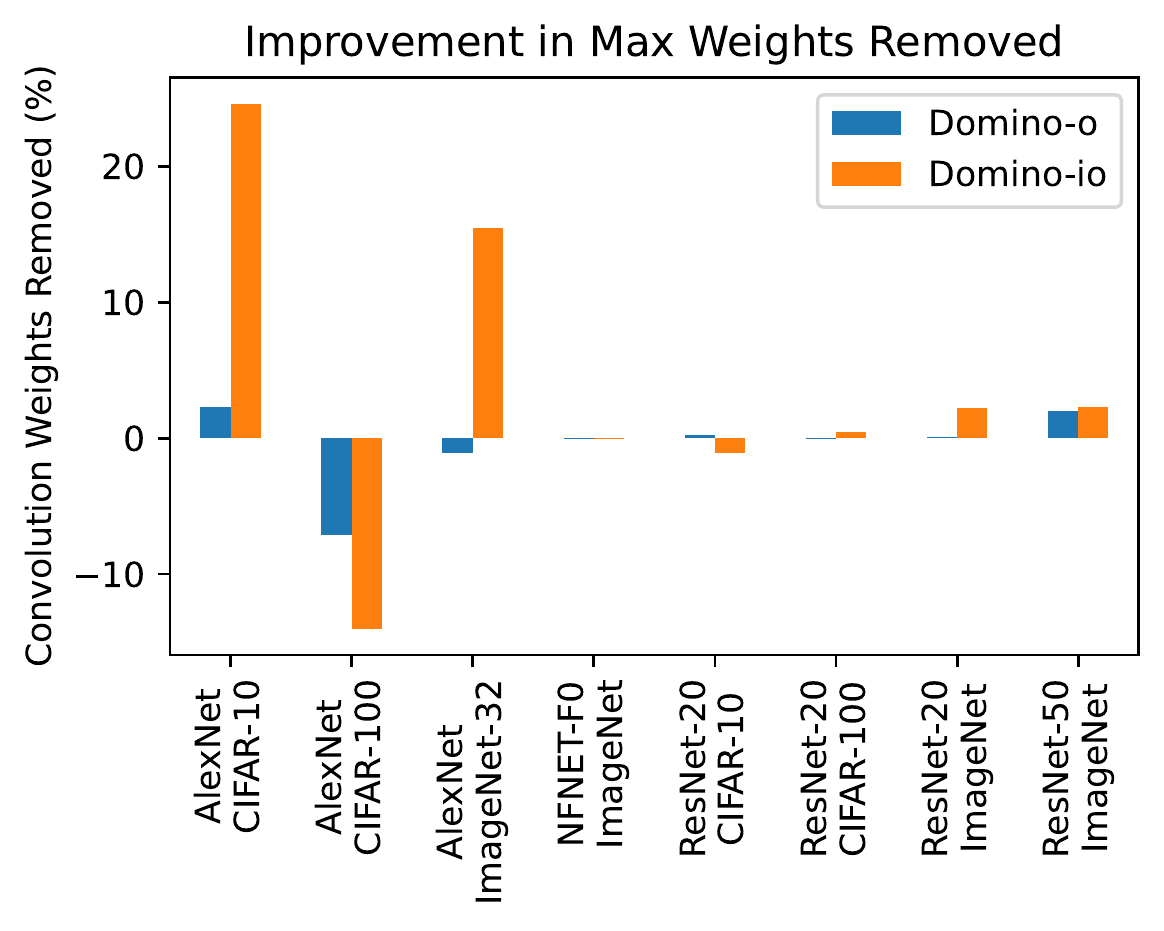}\label{fig:improvement_max}}
\caption{The improvement of domino saliency metrics over the base channel metric.}
\end{figure}

The average improvement of using Domino metrics over channel metrics for each
network, is shown in Figure \ref{fig:improvement}.  On all the networks, except
AlexNet on CIFAR-100, $Domino-io$ on average improves the baseline channel
metric.  $Domino-io$ can be used to push the pruning rate of a network farther.
The improvement shown in Figure \ref{fig:improvement_max} corresponds to the
difference between the maximum percentage of weights removed between the best
Domino metric and the best channel metric for a given network.  With
$Domino-io$, up to 25\% and 15\% more weights can be respectively removed from
AlexNet on the CIFAR-10 dataset and the ImageNet-32 dataset.

\section{Related Work}

When pruning networks with branch connections, multiple channels are 
removed to keep the network dense.  However, most channel pruning approaches
use the same saliency metric \cite{Molchanov2019,Dally} as for simple forward
feed networks.  These saliency metrics do not factor in the other output
channels that need to be removed when one channel is removed.  Some pruning
approaches that are based on feature-map reconstruction
\cite{He2017,ThiNet2019} explicitly describe how branches are taken in account.
He et al.\cite{He2017} and Thinet \cite{ThiNet2019} remove output channels by
considering the effect on the next layer's input feature map.  For the ResNet
architecture, He et al.\cite{He2017} consider the input feature map after the
skip connection. ThiNet \cite{ThiNet2019} avoid pruning layers that could
result in mismatching number of feature maps in the network.   Feature maps
reconstruction-based approaches are computationally expensive.  They are poor
candidates for elaborate pruning schemes
\cite{AMC,SoftFilterPruning,SparseLearningGeneticPruning} that rely on
computationally cost effective saliency metrics.

\section{Conclusion}

Most popular channel saliency metrics do not take into account structural
constraints that can arise when pruning join and split connections.  We propose
two Domino saliency metrics to add structural information to the saliency
measure by combining channel saliency of multiple channels.  $Domino-o$ adds
information about the other output feature maps that are removed together and
$Domino-io$ adds information about output and input feature maps that are
removed together. We observe a small improvement when using $Domino-o$ over the
baseline channel saliency metric and a significant improvement when using
$Domino-io$.  $Domino-io$ can be use to improve the pruning rates by 25\% and
15\% for AlexNet on CIFAR-10 and ImageNet-32.  In conclusion, the use of
$Domino-io$ can significantly improve pruning rates for networks with
join/split connections. This suggests that, in addition to output feature maps
or weights, information about input feature maps or weights is relevant for
pruning.

\section*{Acknowledgement}
This work was supported by Science Foundation Ireland grant 13/RC/2094 to
Lero - The Irish Software Research Centre.

\bibliographystyle{splncs04}
\bibliography{paper}

\end{document}